\documentclass[conference,a4paper]{IEEEtran}
\IEEEoverridecommandlockouts
\usepackage{cite}
\usepackage{algorithmic}
\usepackage{graphicx}
\usepackage{textcomp}
\usepackage{xcolor}
\usepackage{graphicx}
\usepackage{mathrsfs}  
\usepackage{amsfonts,amssymb,amsmath}
\usepackage{hyperref}
\usepackage{multicol,multicap,graphicx,epstopdf}
\usepackage{subcaption}

\DeclareMathOperator*{\argmin}{arg\,min}
\usepackage{booktabs}
\usepackage{multirow}
\usepackage{stmaryrd}
\def\BibTeX{{\rm B\kern-.05em{\sc i\kern-.025em b}\kern-.08em
    T\kern-.1667em\lower.7ex\hbox{E}\kern-.125emX}}
\begin{document}

\title{Correlated Adversarial Joint Discrepancy Adaptation Network
}


\author{\IEEEauthorblockN{Youshan Zhang and Brian D.\ Davison}
\IEEEauthorblockA{\textit{Computer Science and Engineering, Lehigh University, Bethlehem, PA, USA}\\
\{yoz217,\ bdd3\}@lehigh.edu}
}


\maketitle

\IEEEpubidadjcol

\begin{abstract}
Domain adaptation aims to mitigate the domain shift problem when transferring knowledge from one domain into another similar but different domain. However, most existing works rely on extracting marginal features without considering class labels. Moreover, some methods name their model as so-called unsupervised domain adaptation while tuning the parameters using the target domain label. To address these issues, we propose a novel approach called correlated adversarial joint discrepancy adaptation network (\textit{CAJNet}), which minimizes the joint discrepancy of two domains and achieves competitive performance with tuning parameters using the correlated label. By training the joint features, we can align the marginal and conditional distributions between the two domains.  In addition, we introduce a probability-based top-$\mathcal{K}$ correlated label ($\mathcal{K}$-label), which is a powerful indicator of the target domain and effective metric to tune parameters to aid predictions. Extensive experiments on benchmark datasets demonstrate significant improvements in classification accuracy over the state of the art.
\end{abstract}

\begin{IEEEkeywords}
Domain adaptation, Adversarial learning, Distribution alignment
\end{IEEEkeywords}

\section{Introduction}
\label{sec:intro}
The availability of massive labeled training data is a prerequisite of machine learning models. Unfortunately, such a  requirement cannot be met in many real scenarios.
On the other hand, it is time-consuming and expensive to manually annotate data. Therefore, it is often necessary to transfer knowledge from an existing auxiliary labeled domain to a similar but different domain with limited or no labels. However, due to the phenomenon of data bias or domain shift \cite{gong2012geodesic}, machine learning models do not generalize well from an existing domain to a novel unlabeled domain. Domain adaptation (DA) aims to leverage knowledge from an abundant labeled source domain to learn an effective predictor for the target domain with few or no labels, while mitigating the domain shift problem. In this paper, we focus on unsupervised domain adaptation (UDA), where the target domain has no labels. 


Recently, deep neural network methods witness great success in UDA. Especially, adversarial learning shows its power in embedding in deep neural networks to learn feature representations to minimize the discrepancy between the source and target domains~\cite{tzeng2017adversarial,liu2019transferable}. Inspired by the generative adversarial network (GAN)~\cite{goodfellow2014generative}, adversarial learning also contains a feature extractor and a domain discriminator. The domain discriminator aims to distinguish the source domain from the target domain, while the feature extractor aims to learn domain-invariant representations to fool the domain discriminator \cite{ganin2016domain,tzeng2017adversarial,liu2019transferable}. The target domain risk is expected to be minimized via minimax optimization. 

Although many methods achieve remarkable results in domain adaptation, they still suffer from two challenges: (1) the feature extractor often seeks for marginal features without considering the class label information; and (2) it is inappropriate to tune model hyperparameters using real target domain labels, as it violates the setting of UDA. 

To address these challenges, we aggregate four different loss functions in one framework: classification loss, adversarial domain discrepancy loss, top-$\mathcal{K}$ correlated loss, and domain alignment loss to reduce the joint discrepancy of two domains. Moreover, hyperparameters are properly updated using the proposed top-$\mathcal{K}$ correlated label without the target domain label. 
Our contributions are three-fold:
\begin{itemize}
    \item We propose a novel correlated adversarial joint discrepancy adaptation network (\textit{CAJNet}) to adversarially minimize the joint domain discrepancy;
    \item  
    We develop a probabilistic mechanism to compute the joint features for two domains to align both marginal and conditional distributions of source and target domains in a dynamic domain alignment setting;
    \item  We introduce a top-$\mathcal{K}$ correlated loss to help ensure predictions are locally consistent (with those of nearby examples) and use that loss to tune 
    hyperparameters.
\end{itemize} 
Experiments on three benchmark dataset (Office + Caltech-10, Office-31 and Office-Home) 
show that \textit{CAJNet} achieves higher classification accuracy over state-of-the-art methods.

\section{Related Work}

With the advent of GAN~\cite{goodfellow2014generative},  adversarial learning models have been found to be an impactful mechanism for identifying invariant representations in domain adaptation.  The Domain Adversarial Neural Network (DANN) considers a minimax loss to integrate a gradient reversal layer to promote the discrimination of source and target domains~\cite{ganin2016domain}.  The Adversarial Discriminative Domain Adaptation (ADDA) method uses an inverted label GAN loss to split the source and target domain, and features can be learned separately~\cite{tzeng2017adversarial}. The Joint Adaptation Network (JAN)~\cite{long2017deep} combined MMD with adversarial learning to align the joint distribution between two distributions of multiple domain-specific layers across domains. Domain-Symmetric Network (SymNet)~\cite{zhang2019domain} is a symmetrically designed source and target classifier based on an additional classifier. The proposed category level loss can improve the domain level loss by learning the invariant features between two domains.  Miyato et al.~\cite{miyato2018virtual} incorporated virtual adversarial training (VAT) in semi-supervised contexts to smooth the output distributions as a regularization of deep networks. Later, Virtual Adversarial Domain Adaptation (VADA) improved adversarial feature adaptation using VAT. It generated adversarial examples against only the source classifier and adapted on the target domain~\cite{shu2018dirt}. Unlike VADA methods, Transferable Adversarial Training (TAT) adversarially generates transferable examples that fit the gap between source and target domain~\cite{liu2019transferable}.

Through adversarial learning, the domain discrepancy can be largely reduced. Moreover, the feature extractor in deep networks could be large enough to align the feature level (marginal) distribution between domains. However, the features in typical adversarial learning still lack class label information (conditional distribution), \textit{i.e.,} the joint distribution of category level and feature level have not been well addressed.

\section{Method}

\subsection{Problem.}
For unsupervised domain adaptation, given a source domain $\mathcal{D_S} = \{\mathcal{X}_\mathcal{S}^i, \mathcal{Y}_\mathcal{S}^i \}_{i=1}^{\mathcal{N}_\mathcal{S}}$ of $\mathcal{N}_\mathcal{S}$ labeled samples in $C$ categories and a target domain $\mathcal{D_T} = \{\mathcal{X}_\mathcal{T}^j\}_{j=1}^{\mathcal{N}_\mathcal{T}}$ of $\mathcal{N}_\mathcal{T}$ samples without any labels (i.e., $\mathcal{Y}_\mathcal{T}$ is unknown). The samples $\mathcal{X_S}$ and $\mathcal{X_T}$  obey the marginal distribution of $P_{\mathcal{S}}$ and  $P_{\mathcal{T}}$. The conditional distributions of the two domains are denoted as $Q_{\mathcal{S}}$ and $Q_{\mathcal{T}}$. Due to the discrepancy between the two domains, the distributions are assumed to be different, \textit{i.e.,}  $P_{\mathcal{S}} \neq P_{\mathcal{T}}$ and $Q_{\mathcal{S}} \neq Q_{\mathcal{T}}$. 
Our ultimate goal is to learn a classifier $\mathcal{F}$ under a feature extractor $\Phi$,
that ensures lower generalization error in the target domain. 

\begin{figure*}[h]
\centering
\includegraphics[width=1.7\columnwidth]{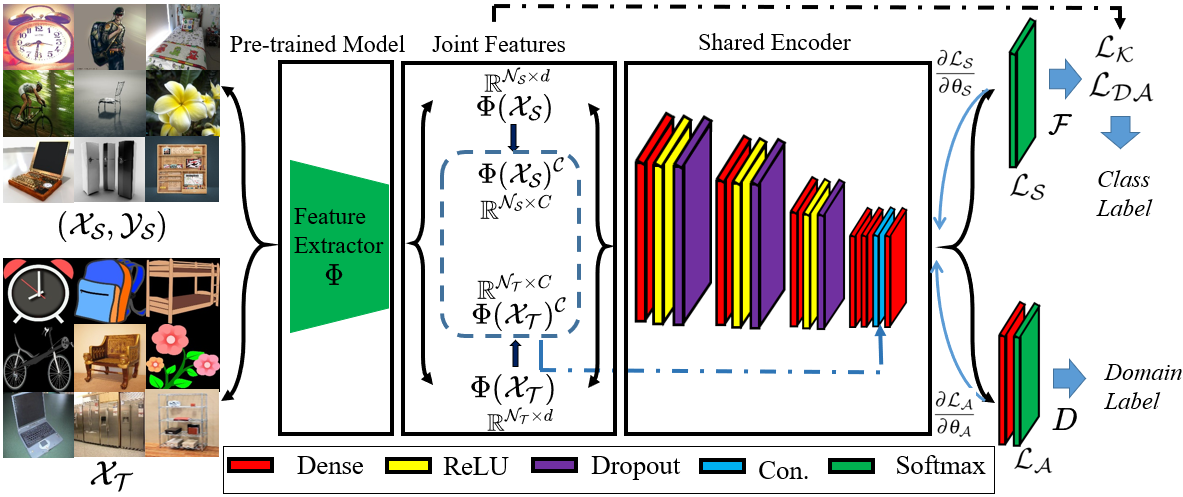}
\caption{The architecture of the \textit{CAJNet} model. We first extract marginal features $\Phi(\mathcal{X_{S/T}}) \in \mathbb{R}^{\mathcal{N}_\mathcal{S/T} \times d}$ for both source and target domains via $\Phi$ using a pre-trained model, and then compute the conditional features $ \Phi(\mathcal{X_{S/T}})^\mathcal{C} \in \mathbb{R}^{\mathcal{N}_\mathcal{S/T} \times C}$ (blue dashed rectangle ), which will be concatenated in the shared encoder (blue dashed line).  The shared encoder is jointly trained with marginal and conditional features. Also, the parameters in the shared encoder are updated by the backward gradients from the class label classifier and the domain label predictor. In addition, the joint features are included in aligning marginal and conditional distributions in $\mathcal{L_{DA}}$ ($\mathcal{L_S}$: classification loss, $\mathcal{L_A}$: adversarial domain loss, $\mathcal{L_K}$: top-$\mathcal{K}$ correlated loss, and $\mathcal{L_{DA}}$: domain alignment loss, Con.: concatenate, $\mathcal{F}$: source classifier and $D$: discriminator, $/$ means or). }
\vspace{-0.3cm}
\label{fig:model}
\end{figure*}

\subsection{Motivation.}
Most existing models were trained only with the extracted marginal features without considering the conditional features. Also, hyperparameters in most deep learning-based models are tuned using target domain accuracy, thus incorrectly exploiting the target labels. A fairer alternative is to develop a metric for tuning to enable impartial assessment of expected performance in the target domain.  Furthermore, most existing work aligns only one distribution (marginal, conditional, or joint) to reduce the domain discrepancy. 

Differently, in this work, we propose first to compute marginal and conditional features and align marginal, conditional, and joint distributions to better align the two domains.  In light of the value of using conditional features, we further develop a top-$\mathcal{K}$ labeling approach for tuning parameters in the target domain.

\subsection{Joint discrepancy in latent space.} 
Considering extracted features are from a pre-trained neural network that maps both domains into a latent space, the marginal discrepancy $ Dist^\mathcal{M}$ between the source and target domain can be formalized as follows.
\begin{equation}\label{eq:dis_m}
    Dist^\mathcal{M}(\mathcal{X_S}, \mathcal{X_T}) = || \frac{1}{\mathcal{N}_\mathcal{S}} \sum_{i=1}^{\mathcal{N}_\mathcal{S}} \Phi(\mathcal{X}_\mathcal{S}^i) - \frac{1}{\mathcal{N}_\mathcal{T}} \sum_{j=1}^{\mathcal{N}_\mathcal{T}} \Phi(\mathcal{X}_\mathcal{T}^j) ||_2,
\end{equation}
 where $||\cdot||_2$ is the L2 norm,  $\Phi ( \cdot) \shortrightarrow \mathbb{R}^d$ is an activation function, which maps samples into a latent $d$-dimensions space, using any pre-trained neural network. Specifically,  $\Phi(x_l) = \Phi (W_{(l)} x_{(l-1)} + b_{(l)} )$, 
where $W_{(l)}, b_{(l)},  x_{(l)}$ are the model weight, bias and output of the $l$-th layer, and then $\Phi(\mathcal{X_{S/T}}) \in \mathbb{R}^{\mathcal{N}_\mathcal{S/T} \times d}$. For the feature extraction, $l$ is the last fully connected layer ($d =1,000$). In the latent space, the marginal distance $Dist^\mathcal{M}$ is minimized via the pre-trained model, and extracted features are frequently used as the single input for traditional models. The conditional discrepancy $Dist^\mathcal{C}$ of two domains should also be minimizing in the following equation.
\vspace{-0.3cm}
\begin{equation}\label{eq:dis_c}
\begin{aligned}
    Dist^\mathcal{C}&(\mathcal{Y_S}\vert \mathcal{X_S}, \mathcal{Y_T }\vert \mathcal{X_T})  = Dist^\mathcal{C} \sum_{c=1}^{C}(\mathcal{Y_{S^\textit{c}}}\vert \mathcal{X_{S^\textit{c}}}, \mathcal{Y_{T^\textit{c}} }\vert \mathcal{X_{T^\textit{c}}}) \\ & 
    = \sum_{c=1}^{C} || \frac{1}{\mathcal{N}_\mathcal{S}^{c}} \sum_{i=1}^{\mathcal{N}_\mathcal{S}^c} \Phi(\mathcal{X}_\mathcal{S^\textit{c}}^i) - \frac{1}{\mathcal{N}_\mathcal{T}^c} \sum_{j=1}^{\mathcal{N}_\mathcal{T}^c} \Phi(\mathcal{X}^j_\mathcal{T^\textit{c}}) ||_2,
\end{aligned}
\end{equation}
where $C$ is the number of categories, $\mathcal{Y_{S^\textit{c}}}\vert \mathcal{X_{S^\textit{c}}}$ (or $\mathcal{Y_{T^\textit{c}}}\vert \mathcal{X_{T^\textit{c}}}$)  represents $c^{th}$ category target (sub-domain) data in the source (or target) domain and $\mathcal{N}_\mathcal{S}^c$ (or $\mathcal{N}_\mathcal{T}^c$) is the number of samples in the  $c^{th}$ category in the source or target domain. However, we cannot directly compute such a  conditional distance due to the unknown $\mathcal{Y_T }$ or $\Phi(\mathcal{X}^j_\mathcal{T^\textit{c}})$. We hence need to reformulate the sub-domain data, so that it contains the $C$ categories' label information. In the latent space, if we only consider the marginal discrepancy, supposing that $\mathcal{Y}_\mathcal{T}^j = c$, the probability of marking a correct prediction of $\mathcal{X}_\mathcal{T}^j$ is
\begin{equation}\label{eq:pro}
\begin{aligned}
p_{j}^{\mathcal{M}} & = \sum_{i=1}^{\mathcal{N}_\mathcal{S}^c} p_{j\vert i}^{\mathcal{M}}  = \sum_{i=1}^{\mathcal{N}_\mathcal{S}^c} \frac{Dist^\mathcal{M}(\mathcal{X}_\mathcal{S}^i, \mathcal{X}_\mathcal{T}^j)}{\sum_{i=1}^{\mathcal{N}_\mathcal{S}}Dist^\mathcal{M}(\mathcal{X}_\mathcal{S}^i, \mathcal{X}_\mathcal{T}^j)} \\ &= \frac{\sum_{i=1}^{\mathcal{N}_\mathcal{S}^c} ||\Phi(\mathcal{X}_\mathcal{S}^i) - \Phi(\mathcal{X}_\mathcal{T}^j) ||_2}{\sum_{i=1}^{\mathcal{N}_\mathcal{S}} || \Phi(\mathcal{X}_\mathcal{S}^i)- \Phi(\mathcal{X}_\mathcal{T}^j) ||_2}.
\end{aligned}
\end{equation}
Motivated thus, we model the target conditional features:
\begin{equation}\label{eq:con}
\begin{aligned}
 & \Phi(\mathcal{{X_{T}})^\mathcal{C}} = \mathcal{Y_{T^\textit{c}} }\vert \mathcal{X_{T^\textit{c}}} \\ &\approx \otimes_{j=1}^{\mathcal{N}_\mathcal{T}}\oplus_{c=1} ^C || \frac{1}{\mathcal{N}_\mathcal{S}^{c}} \sum_{i=1}^{\mathcal{N}_\mathcal{S}^c} \Phi(\mathcal{X}_\mathcal{S}^i)- \Phi(\mathcal{X}_\mathcal{T}^j) ||_2,
\end{aligned}
 \end{equation}
where $\otimes$ and $\oplus$ is the column and row matrix concatenation operation, respectively. One target sample has the $C-$dimensions and  $ \Phi(\mathcal{{X_{T}})^\mathcal{C}} \in \mathbb{R}^{\mathcal{N}_\mathcal{T} \times C}$. Similarly, the source conditional features can be denoted as $\Phi(\mathcal{{X_{S}})^\mathcal{C}} = \mathcal{Y_{S^\textit{c}} }\vert \mathcal{X_{S^\textit{c}}} \approx \otimes_{i=1}^{\mathcal{N}_\mathcal{S}}\oplus_{c=1} ^C || \frac{1}{\mathcal{N}_\mathcal{S}^{c}} \sum_{i=1}^{\mathcal{N}_\mathcal{S}^c} \Phi(\mathcal{X}_\mathcal{S}^i)- \Phi(\mathcal{X}_\mathcal{S}^{i'}) ||_2 \in \mathbb{R}^{\mathcal{N}_\mathcal{S} \times C}$ ($\mathcal{X}_\mathcal{S}^{i'}$ is one element in $\mathcal{X}_\mathcal{S}$). Therefore, we rewrite Eq.~\ref{eq:dis_c} as follows.
%
\begin{equation}\label{eq:cond}
\begin{aligned}
    &Dist^\mathcal{C}(\mathcal{Y_S}\vert \mathcal{X_S}, \mathcal{Y_T }\vert \mathcal{X_T}) \approx Dist^\mathcal{C}(\Phi(\mathcal{{X_{S}})^\mathcal{C}}, \Phi(\mathcal{{X_{T}})^\mathcal{C}}) \\ & = || \frac{1}{\mathcal{N}_\mathcal{S}} \sum_{i=1}^{\mathcal{N}_\mathcal{S}} \Phi(\mathcal{X}_\mathcal{S}^i)^\mathcal{C}- \frac{1}{\mathcal{N}_\mathcal{T}} \sum_{j=1}^{\mathcal{N}_\mathcal{T}} \Phi(\mathcal{X}_\mathcal{T}^j)^\mathcal{C} ||_2
\end{aligned}
 \end{equation}
If we only consider the conditional discrepancy, the probability of making correct prediction of $\mathcal{X}_\mathcal{T}^j$ is denoted as:
\begin{equation}\label{eq:con_pro}
\begin{aligned}
p_{j}^{\mathcal{C}} & = \sum_{i=1}^{\mathcal{N}_\mathcal{S}^c} p_{j\vert i}^{\mathcal{C}}  = \sum_{i=1}^{\mathcal{N}_\mathcal{S}^c} \frac{Dist^\mathcal{C}(\mathcal{Y}_\mathcal{S}^i \vert \mathcal{X}_\mathcal{S}^i, \mathcal{Y}_\mathcal{T}^j \vert \mathcal{X}_\mathcal{T}^j)}{\sum_{i=1}^{\mathcal{N}_\mathcal{S}}Dist^\mathcal{C}(\mathcal{Y}_\mathcal{S}^i \vert \mathcal{X}_\mathcal{S}^i, \mathcal{Y}_\mathcal{T}^j \vert \mathcal{X}_\mathcal{T}^j)} \\ & = \frac{\sum_{i=1}^{\mathcal{N}_\mathcal{S}^c}|| \Phi(\mathcal{X}_\mathcal{S}^i)^\mathcal{C}- \Phi(\mathcal{X}_\mathcal{T}^j)^\mathcal{C} ||_2}{\sum_{i=1}^{\mathcal{N}_\mathcal{S}}||\Phi(\mathcal{X}_\mathcal{S}^i)^\mathcal{C}- \Phi(\mathcal{X}_\mathcal{T}^j)^\mathcal{C} ||_2}.
\end{aligned}
\end{equation}
Substituting Eq.~\ref{eq:dis_m} and Eq.~\ref{eq:cond}, we define the \textbf{joint discrepancy} in the latent space in Eq.~\ref{eq:joint}.
\begin{equation}\label{eq:joint}
\begin{aligned}
    Dist(\mathcal{D_S}, \mathcal{D_T}) & = Dist^\mathcal{M}(\mathcal{X_S}, \mathcal{X_T}) \\ & +Dist^\mathcal{C}(\Phi(\mathcal{{X_{S}})^\mathcal{C}}, \Phi(\mathcal{{X_{T}})^\mathcal{C}}) 
\end{aligned}
\end{equation}
The \textbf{joint features} of the source and target domain are defined as $\mathcal{J}(\mathcal{X_{S}}) = \Phi(\mathcal{X_{S}})\odot  \Phi(\mathcal{{X_{S}})^\mathcal{C}}$ and $\mathcal{J}(\mathcal{X_{T}}) = \Phi(\mathcal{X_{T}})\odot  \Phi(\mathcal{{X_{T}})^\mathcal{C}}$, where $\odot$ is the feature conjunct function. 

Substituting Eq.~\ref{eq:pro} and Eq.~\ref{eq:con_pro}, the probability of making correct prediction of $\mathcal{X}_\mathcal{T}^j$ considering both marginal and conditional discrepancies is given by: 
\vspace{-0.3cm}
\begin{equation}\label{eq:pro_mc}
\begin{aligned}
&p_{j}^{\mathcal{M, C}} = \sum_{i=1}^{\mathcal{N}_\mathcal{S}^c} (p_{j\vert i}^{\mathcal{M}} \circ p_{j\vert i}^{\mathcal{C}})  \\& = \frac{\sum_{i=1}^{\mathcal{N}_\mathcal{S}^c} (|| \Phi(\mathcal{X}_\mathcal{S}^i)-\Phi(\mathcal{X}_\mathcal{T}^j) ||_2+ || \Phi(\mathcal{X}_\mathcal{S}^i)^\mathcal{C}- \Phi(\mathcal{X}_\mathcal{T}^j)^\mathcal{C} ||_2)}{\sum_{i=1}^{\mathcal{N}_\mathcal{S}}(|| \Phi(\mathcal{X}_\mathcal{S}^i)- \Phi(\mathcal{X}_\mathcal{T}^j) ||_2+ ||  \Phi(\mathcal{X}_\mathcal{S}^i)^\mathcal{C}- \Phi(\mathcal{X}_\mathcal{T}^j)^\mathcal{C} ||_2)}
\end{aligned}
\end{equation}
where $\circ$ is the probability conjunct function. 

Since we want to minimize the joint discrepancy of the two domains in Eq.~\ref{eq:joint}, it is equivalent to maximize  $\sum_{j=1}^{\mathcal{N_T}}p_{j}^{\mathcal{M, C}}$ in the target domain. To realize this, our proposed \textit{CAJNet} optimizes a four-component loss function using the joint features, as shown in Fig.~\ref{fig:model}. In summary, the discrepancies of both the marginal and conditional distributions between the source and target
can be jointly reduced by minimizing $Dist(\mathcal{D_S}, \mathcal{D_T})$. Specifically, the discrepancy between $P_{\mathcal{S}}$ and  $P_{\mathcal{T}}$ can be reduced by minimizing $Dist^\mathcal{M}$, whereas the mismatches of $Q_{\mathcal{S}}$ and  $Q_{\mathcal{T}}$ can be decreased by minimizing $Dist^\mathcal{C}$.

\begin{figure}[hb!]
    \centering
    \includegraphics[width=.9\columnwidth]{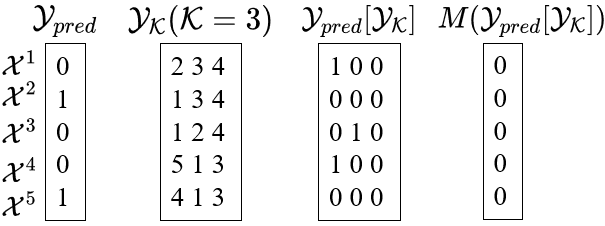}
    \caption{An example of top-$\mathcal{K}$ (here, 3) correlated labels in updating predicted labels. Given five examples ($\mathcal{X}^1$ to $\mathcal{X}^5$), the prediction is $\mathcal{Y}_{pred}$. The predictions of the five examples contain two classes, 
    while the true labels are all zeros. There are two incorrectly predicted results ($\mathcal{Y}_{pred}^2$ and $\mathcal{Y}_{pred}^5$). The top-3 $\mathcal{Y}_{\mathcal{K}}$ shows the top 3 instances that should be in the same class (e.g., $\mathcal{X}^1$ should have the same class as $\mathcal{X}^2$, $\mathcal{X}^3$ and $\mathcal{X}^4$). The $M(\mathcal{Y}_{pred}[\mathcal{Y}_{\mathcal{K}}])$ changes the predicted labels and (in this case) outputs the same label as the truth label.}
    \label{fig:top_k}
    \vspace{-0.3cm}
\end{figure}

\subsection{Top $\mathcal{K}$ correlated label ($\mathcal{K}$-label).}

Although there are no labels available in the target domain,  model parameters still need to be tuned based on target domain information for pursuing higher transfer accuracy. We hence calculate the top-$\mathcal{K}$ correlated labels for the target domain. Eq.~\ref{eq:pro_mc} considers the probability of cross domains, while in the same domain, we can also denote a similar probability $p_{j|n}$ to indicate whether $\mathcal{X}_\mathcal{T}^j$ would pick $\mathcal{X}_\mathcal{T}^n$ ($n \neq j$) as its neighbor. For nearby data points, $p_{j|n}$ is relatively high, whereas for widely separated data points, $p_{j|n}$ will be almost infinitesimal.

The probabilistic similarity between $\mathcal{X}_\mathcal{T}^j$ and $\mathcal{X}_\mathcal{T}^n$ is defined:
\begin{equation*}\label{eq:pro_c}
\begin{aligned}
& p_{j|n}^{\mathcal{M, C}}  = p_{j\vert n}^{\mathcal{M}} \circ p_{j\vert n}^{\mathcal{C}}  \\ & = \frac{ ||  \Phi(\mathcal{X}_\mathcal{T}^n)- \Phi(\mathcal{X}_\mathcal{T}^j)||_2+ ||  \Phi(\mathcal{X}_\mathcal{T}^n)^\mathcal{C}- \Phi(\mathcal{X}_\mathcal{T}^j)^\mathcal{C} ||_2}{\sum_{n=1}^{\mathcal{N}_\mathcal{T}}(||  \Phi(\mathcal{X}_\mathcal{T}^n)- \Phi(\mathcal{X}_\mathcal{T}^j) ||_2+ ||  \Phi(\mathcal{X}_\mathcal{T}^n)^\mathcal{C}- \Phi(\mathcal{X}_\mathcal{T}^j)^\mathcal{C} ||_2)}.
\end{aligned}
\end{equation*}
This probability matrix in the target domain is $P_{\mathcal{T}_m} =\otimes_{j=1}^{\mathcal{N}_\mathcal{T}}\oplus_{n=1} ^{\mathcal{N}_\mathcal{T}} p_{j|n}^{\mathcal{M, C}}$ ($p_{j|j}^{\mathcal{M, C}} = 0$). Hence, $P_{\mathcal{T}_m} \in \mathbb{R}^{\mathcal{N}_\mathcal{T} \times \mathcal{N}_\mathcal{T}}$ with its elements $0 \leq P_{jn} \leq 1$, also has the constraint of $\sum_{n=1}^{\mathcal{N}_\mathcal{T}} P_{jn} =1, \ \forall n \in \{1, \cdots, \mathcal{N}_\mathcal{T}\}$. The $\mathcal{K}$-label measures the index of the most similar sample in the same domain, which is reformulated as $\mathcal{Y}_\mathcal{T_K} = \otimes_{j=1}^{\mathcal{N}_\mathcal{T}} \uparrow (P_{\mathcal{T}_m}^{j})\in \mathbb{R}^{\mathcal{N_T}\times \mathcal{K}}, \ \mathcal{Y}_\mathcal{S_K} = \otimes_{i=1}^{\mathcal{N}_\mathcal{S}} \uparrow (P_{\mathcal{S}_m}^{i}) \in \mathbb{R}^{\mathcal{N_S}\times \mathcal{K}}$,
%
where $\uparrow$ is an operation that first descending sort
the probability matrix $P_{\mathcal{T}_m}^{j}/P_{\mathcal{S}_m}^{i}$ and then return the first $\mathcal{K}$ index of associated probability. The defined top $\mathcal{K}$-label contains the $k$ nearest neighborhood, that has the top  $k$ highest probability for selecting one sample in the same category. Therefore, the top-$\mathcal{K}$ correlated loss in the target domain is defined in Eq.~\ref{eq:lk}.
\vspace{-0.1cm}
\begin{equation}
\begin{aligned}\label{eq:lk}
    \mathcal{L_K}(\mathcal{Y}_{pred}, \mathcal{Y}_{\mathcal{K}})=\frac{1}{\mathcal{N}_\mathcal{S/T}}\sum_{i/j=1}^{\mathcal{N}_\mathcal{S/T}}  |\mathcal{Y}_{pred}^{i/j} - M(\mathcal{Y}_{pred}[\mathcal{Y}_{\mathcal{K}}^{i/j}])|
 \end{aligned}
\end{equation}
where $|\cdot|$ is the absolute value operation, $\mathcal{Y}_{pred}^{i/j}$ is the prediction of either source domain or target domain from source classifier in Eq.~\ref{eq:lc}, and $\mathcal{Y}_{\mathcal{K}}^{i/j}$ is the correlated label with size $\mathcal{N_{S/T}\times \mathcal{K}}$; it shows the top-$\mathcal{K}$ index, which is highly related to the instance that should be in the same class. $\mathcal{Y}_{pred}[{\mathcal{Y_K}}]$ is the updated matrix for the predicted labels and $M (\cdot) $ selects the most frequent labels in the updated matrix as shown in Fig.~\ref{fig:top_k}. Therefore, the loss measures how different the predicted label is to its nearest neighbors, which are assumed to be from the same domain.
\subsection{Source classifier.}
The task in the source domain is trained using the typical cross-entropy loss:
\begin{equation}\label{eq:lc}
    \mathcal{L_S} (\mathcal{F}(\mathcal{J}(\mathcal{X_{S}})),\mathcal{Y_{S}}) = - \frac{1}{\mathcal{N}_\mathcal{S}}\sum_{i=1}^{\mathcal{N}_\mathcal{S}} \sum_{c=1}^{C} \mathcal{Y}_{\mathcal{S}_{c}}^{i} \text{log}  (\mathcal{F}(\mathcal{J}(\mathcal{X}_{\mathcal{S}_{c}}^{i}))), 
\end{equation}
where $\mathcal{Y}_{\mathcal{S}_{c}}^{i} \in [0, 1]^{C}$ is the binary indicator of each class $c$ in true label, 
and $\mathcal{F}(\mathcal{J}(\mathcal{X}_{\mathcal{S}_{c}}^{i}))$ is the predicted probability of class $c$ using classifier $\mathcal{F}$.
\subsection{Adversarial domain loss.}
Given the feature representation output of shared encoder, we can learn a discriminator $D$ as shown in Fig.~\ref{fig:model}, which can distinguish the two domains using following loss function:
\begin{gather}\label{eq:all1}
\scalebox{0.99}{$
\begin{aligned}
    \mathcal{L_A}(\mathcal{J}(\mathcal{X_{S}}), \mathcal{J}(\mathcal{X_{T}}))   =  &- \frac{1}{\mathcal{N_S}} \sum_{i=1}^{\mathcal{N_S}} \text{log} (1-D(\mathcal{J}(\mathcal{X}_\mathcal{S}^i))) \\ & - \frac{1}{\mathcal{N_T}} \sum_{j=1}^{\mathcal{N_T}} \text{log} (D(\mathcal{J}(\mathcal{X}_\mathcal{T}^j))) 
\end{aligned}$}
\end{gather}

\subsection{Shared Encoder.}
The shared encoder begins with three repeated blocks and each block has a dense layer, a ``ReLU" activation layer, and a dropout layer. The numbers of units of the dense layer are 512, 128, and 64, respectively. The rate of the Dropout layer is 0.5. It ends with a dense layer (the number of units is the number of classes in each dataset). The shared encoder is jointly optimized by both the source classifier and the domain labels. Note that the input of shared encoder is the joint features: the marginal features $\Phi(\mathcal{X_S})$ are first fed into the repeated blocks and then the conditional features $\Phi(\mathcal{X_S}^{\mathcal{C}})$ are concatenated.


Let $\mathcal{F_E} ( \cdot,\theta_\mathcal{E})$ be the output of shared encoder with parameters of $\theta_\mathcal{E}$. In addition, let $\mathcal{F_S} ( \cdot,\theta_\mathcal{S})$ be output of class label classifier with parameters of $\theta_\mathcal{S}$ and $\mathcal{F_A} ( \cdot,\theta_\mathcal{A})$ be output of domain label predictor with parameters of $\theta_\mathcal{A}$. Therefore, the shared encoder is optimized by both class label classifier and domain label predictor. The parameters are updated as:
%
\begin{equation}
\begin{aligned}\label{eq:share}
& \theta_\mathcal{S} \shortleftarrow \theta_\mathcal{S} - \epsilon\frac{\partial \mathcal{L_S}}{\partial \theta_\mathcal{S} }, \ \theta_\mathcal{A} \shortleftarrow \epsilon\tau\theta_\mathcal{A} \frac{\partial \mathcal{L_A}}{\partial \theta_\mathcal{A} },  \ \\ & \theta_\mathcal{E} \shortleftarrow \theta_\mathcal{E} - \epsilon (\frac{\partial \mathcal{L_S}}{\partial \theta_\mathcal{S}} -\tau\theta_\mathcal{A} \frac{\partial \mathcal{L_A}}{\partial \theta_\mathcal{A} }), 
\end{aligned}
\end{equation}
where $\epsilon$ is learning rate and $\tau$ is the adaptation factor.

\subsection{Dynamic distribution alignment.}
\vspace{-0.1cm}
We can get the prediction of the target domain from the class label classifier.  However, we can further improve the predicted accuracy by employing a dynamic distribution alignment, which can dynamically balance the marginal and conditional distribution and update the predicted labels in the target domain. Manifold Embedded Distribution Alignment (MEDA), proposed by Wang et al.~\cite{wang2018visual}, aligns learned features from manifold learning. However, Zhang et al.\ showed that there are defects in the MEDA model, which cannot estimate the geodesic of sub-source and sub-target domains \cite{zhang2019transductive}.  We modified the domain alignment loss as follows:
\begin{equation*}
\begin{aligned}\label{eq:meda}
\mathcal{L_{DA}}(\mathcal{D_S}, \mathcal{D_T})=&\mathop{\arg\min}  \mathcal{L_S} (\mathcal{F} (\mathcal{J} (\mathcal{X_{S}})),\mathcal{Y_{S}}) +\eta ||\mathcal{F} ||_{K}^{2} \\ & +\lambda \overline{D_{\mathcal{F} }} (\mathcal{D_S},\mathcal{D_T} ) + \rho R_{\mathcal{F} }(\mathcal{D_S},\mathcal{D_T})
\end{aligned}
\end{equation*}
where $\mathcal{F}$ is the classifier from the shared encoder, $\mathcal{L_{G}}$ is the sum of squares loss; $||\mathcal{F} ||^2$ is the squared norm of $\mathcal{F}$; and the first two terms minimize the structure risk of shared encoder.  $\overline{D_{\mathcal{F} }}(\cdot, \cdot)$ represents the dynamic distribution alignment; $R_{\mathcal{F} }(\cdot, \cdot)$ is a Laplacian regularization; $\eta, \lambda$, and $\rho$ are regularization parameters. Specifically, $\overline{D_{\mathcal{F} }} (\mathcal{D_S},\mathcal{D_T} )= (1-\mu)D_{\mathcal{F} } (P_\mathcal{S}, P_\mathcal{T}) + \mu \sum_{c=1}^{C}D_{\mathcal{F} }^{c} (Q_\mathcal{S}, Q_\mathcal{T})$, where $\mu$ is an adaptive factor to balance the marginal distribution $(P_\mathcal{S}, P_\mathcal{T})$, and conditional distribution $(Q_\mathcal{S}, Q_{\mathcal{T}})$ \cite{wang2018visual}.

\subsection{{\em CAJNet} model.}
The framework diagram of our proposed \textit{CAJNet} model is depicted in Fig.~\ref{fig:model}.  Our model minimizes the following objective function:
\begin{equation}
\begin{aligned}\label{eq:all}
& \mathcal{L}  (\mathcal{X_S}, \mathcal{Y_S},  \mathcal{X_T}, \mathcal{Y_{T_K}}) =  \mathop{\argmin}  (\mathcal{L_S} (\mathcal{F}(\mathcal{J}(\mathcal{X_{S}})), \mathcal{Y_{S}})  \\ &   + \mathcal{L_A}(\mathcal{J}(\mathcal{X_{S}}), \mathcal{J}(\mathcal{X_{T}}))   + \mathcal{L_K}(\mathcal{Y}_{pred}, \mathcal{Y}_{\mathcal{T_K} })   + \mathcal{L_{DA}}(\mathcal{D_S}, \mathcal{D_T}))
\end{aligned}
\end{equation}
where $\mathcal{F}$ is the source classifier in Eq.~\ref{eq:lc}; $\mathcal{L_S}$ is the cross-entropy loss; $\mathcal{Y}_{pred}$ is the predicted label and $\mathcal{Y_{T_K}}$ is the top-$\mathcal{K}$ correlated label, which shows the $\mathcal{K}$ most highly related samples. $\mathcal{J}(\mathcal{X_{S}})$ and $\mathcal{J}(\mathcal{X_{T}})$ are the joint features. The $\mathcal{L_A}$, $\mathcal{L_K}$, and $\mathcal{L_{DA}}$ represent the adversarial domain loss, top-$\mathcal{K}$ correlated loss, and the domain alignment loss. 

In summary, to effectively align both 
marginal and conditional distributions, we integrate the aforementioned four different loss functions in our proposed \textit{CAJNet} model. The correlated adversarial learning mechanism further improves the domain invariant and leads to better results.  The performance of different loss functions is shown in Sec.~\ref{sec:abl}. 

\section{Experiments}

\subsection{Datasets.}
\textbf{Office + Caltech-10} \cite{gong2012geodesic} consists of Office~10 and Caltech~10 datasets with 2,533 images in 10 classes of four domains: Amazon (A), Webcam (W), DSLR (D) and Caltech (C). 
C$\shortrightarrow$A represents learning knowledge from domain C which is applied to domain A. \textbf{Office-31} \cite{saenko2010adapting} consists of 4,110 images in 31 classes from three  domains: Amazon (A), Webcam (W), and DSLR (D). \textbf{Office-Home} \cite{venkateswara2017deep} contains 15,588 images from 65 categories. It has four domains: Art (Ar), Clipart (Cl), Product (Pr) and Real-World (Rw).

\subsection{Implementation details.}
We implement our approach using PyTorch with an Nvidia GeForce 1080 Ti GPU and extract features for the three datasets from a fine-tuned ResNet50 network~\cite{he2016deep}. The 1,000 features are then extracted from the last fully connected layer~\cite{zhang2020impact}. Parameters in domain distribution alignment are  $\eta = 0.1$, $\lambda=10$, and $\rho= 10$, which are fixed based on previous research~\cite{wang2018visual}, and $\tau = 0.31$ is from~\cite{ghifary2014domain}. Learning rate ($\epsilon = 0.001$), batch size (32), and number of epochs (1000) are determined by the performance of source domain.   $\mathcal{K} = 3$ is tuned by minimizing $\mathcal{L_K}$. We compare our results with 11 state of the art (including traditional methods and deep networks).

\begin{table*}[t]
\begin{center}
\caption{Accuracy (\%) on Office + Caltech-10 dataset}
\vspace{-0.2cm}
\setlength{\tabcolsep}{+2.3mm}{
\begin{tabular}{rccccccccccccc}
\hline \label{tab:OC+10}
Task & C$\shortrightarrow$A &  C$\shortrightarrow$W & C$\shortrightarrow$D & A$\shortrightarrow$C & A$\shortrightarrow$W & A$\shortrightarrow$D & W$\shortrightarrow$C & W$\shortrightarrow$A & W$\shortrightarrow$D & D$\shortrightarrow$C & D$\shortrightarrow$A & D$\shortrightarrow$W & \textbf{Ave.}\\
\hline
\textbf{JDA}~\cite{long2013transfer}&95.3&   96.3&   96.8&   93.9&   95.9&   95.5&   93.5&   95.7&  \textbf{100} & 93.3&   95.5&   96.9&   95.7\\
\textbf{CORAL}~\cite{sun2017correlation}	& 95.6&   96.3&   98.1&   95.2&   89.8&   94.3&   93.9&   95.7&  \textbf{100} &  94.0 & 96.2&   98.6&   95.6\\
\textbf{MEDA}~\cite{wang2018visual}	& 96.0 & 99.3 &   98.1&   94.2&   99.0 &\textbf{100} & 94.6 &   96.5 &  \textbf{100} &  94.1&   96.1&   99.3&   97.3\\
\hline
\hline
RTN~\cite{long2016unsupervised} &93.7 & 96.9 &94.2 &88.1 &95.2 & 95.5& 86.6& 92.5& \textbf{100} & 84.6& 93.8 & 99.2 &93.4 \\
MDDA~\cite{rahman2020minimum} &93.6 & 95.2 &93.4 &89.1 &95.7 & 96.6& 86.5&94.8 & \textbf{100} & 84.7& 94.7 & 99.4 & 93.6\\
SSD~\cite{zhang2020domain} &\textbf{96.2}&	96.2&	96.2&	\textbf{94.9}&	98.3&	98.3&	\textbf{98.1} &\textbf{98.1}	&98.7&	\textbf{96.2}&	96.2&	99.3&	97.2 \\
 \hline
  \hline
 \textbf{\textit{CAJNet}}& \textbf{96.2} & \textbf{100} &   \textbf{100} &   94.6 & \textbf{  100} & \textbf{100} & 94.2 &   96.7 &  \textbf{100} &  94.6 &   \textbf{96.7}&   \textbf{100} &   \textbf{97.8}\\
  \hline
\end{tabular}}
\vspace{-0.3cm}
\end{center}
\end{table*}

\begin{table*}[h!]
\begin{center}
\caption{Accuracy (\%) on Office-Home dataset}
\vspace{-0.2cm}
\setlength{\tabcolsep}{+1.6mm}{
\begin{tabular}{rccccccccccccc}
\hline \label{tab:OH}
Task & Ar$\shortrightarrow$Cl &  Ar$\shortrightarrow$Pr & Ar$\shortrightarrow$Rw & Cl$\shortrightarrow$Ar & Cl$\shortrightarrow$Pr & Cl$\shortrightarrow$Rw & Pr$\shortrightarrow$Ar & Pr$\shortrightarrow$Cl & Pr$\shortrightarrow$Rw & Rw$\shortrightarrow$Ar & Rw$\shortrightarrow$Cl & Rw$\shortrightarrow$Pr & \textbf{Ave.}\\
\hline
\textbf{JDA}~\cite{long2013transfer}& 47.4&	72.8&	76.1&	60.7&	68.6&	70.5&	66.0 &	49.1&	76.4&	69.6&	52.5&	79.7&	65.8 \\
\textbf{CORAL}~\cite{sun2017correlation}	& 48.0 &	78.7&	80.9&	65.7&	74.7&	75.5&	68.4&	49.8&	80.7&	73.0	&50.1&	82.4&	69.0\\
\textbf{MEDA}~\cite{wang2018visual}& 48.5	&74.5&	78.8&	64.8&	76.1&	75.2&	67.4&	49.1&	79.7&	72.2&	51.7&	81.5	&68.3\\
\hline
\hline
DANN~\cite{ghifary2014domain}	& 45.6	& 59.3& 	70.1& 	47.0& 	58.5& 	60.9& 	46.1& 	43.7& 	68.5& 	63.2& 	51.8& 	76.8& 	57.6\\
JAN~\cite{long2017deep}	& 45.9& 	61.2& 	68.9& 	50.4& 	59.7& 	61.0& 	45.8& 	43.4& 	70.3& 	63.9& 	52.4& 	76.8& 	58.3\\
TADA~\cite{wang2019transferable} & 53.1 &72.3& 77.2& 59.1 &71.2& 72.1& 59.7& 53.1& 78.4 &72.4& 60.0 &82.9& 67.6 \\
SymNets~\cite{zhang2019domain}& 47.7 & 72.9 & 78.5 & 64.2  & 71.3  &74.2  & 64.2  & 48.8 &  79.5&  74.5 &52.6 & 82.7& 67.6 \\
\hline
\hline
\textbf{\textit{CAJNet}}& \textbf{58.0} &	\textbf{82.9} &	\textbf{83.9} &	\textbf{74.5} &	\textbf{84.2} &	\textbf{83.7} &	\textbf{74.8}&	\textbf{57.9} &	\textbf{85.0} &	\textbf{76.6} &	\textbf{60.1} &	\textbf{87.4} & 	\textbf{75.8}\\
\hline
\end{tabular}}
\end{center}
\vspace{-0.6cm}
\end{table*}

\begin{table}[!htb]
      \caption{Accuracy (\%) on Office-31 dataset}
      \vspace{-0.2cm}
      \centering
\setlength{\tabcolsep}{+0.9mm}{
\begin{tabular}{rcccccccc|c|c|c|c|c|c|c|c|}
\hline \label{tab:O31}
Task & A$\shortrightarrow$W &  A$\shortrightarrow$D & W$\shortrightarrow$A & W$\shortrightarrow$D & D$\shortrightarrow$A & D$\shortrightarrow$W  & \textbf{Ave.}\\
\hline
\textbf{JDA}~\cite{long2013transfer}& 79.1 &  79.7  & 72.9 &  97.4  & 71.0  & 94.2 &  82.4\\
\textbf{CORAL}~\cite{sun2017correlation}	& 88.9 &  87.6 &  74.7 &  99.2 &  73.0  & 96.7 &  86.7 \\
\textbf{MEDA}~\cite{wang2018visual}	& 90.8 &  91.4 &  74.6  & 97.2 &  75.4  & 96.0 &  87.6\\
\hline
\hline
JAN~\cite{long2017deep}&	85.4 &	84.7	&70.0 &	99.8	&68.6 &	97.4 &	84.3\\
TADA~\cite{wang2019transferable} &94.3 & 91.6  & 73.0  &99.8  & 72.9 & 98.7 & 88.4 \\
SymNets~\cite{zhang2019domain}& 90.8& 93.9  &72.5  & \textbf{100} & 74.6 & 98.8& 88.4 \\
CAN~\cite{kang2019contrastive} & 94.5 & 95.0  &77.0   &99.8  & 78.0 & \textbf{99.1} &  90.6 \\
\hline
\hline
\textbf{$CAJNet-\mathcal{M}$}& 93.7  & 93.0  & 78.5 &  98.5  & 80.1  & 97.9 & 90.3\\
\textbf{$CAJNet-\mathcal{C}$}& 95.0  & 95.3  & 79.7 &  98.8  & 79.9  & 98.0 & 91.1\\
\textbf{\textit{CAJNet}}& \textbf{96.0}  & \textbf{95.6}  & \textbf{79.9} &  99.2  & \textbf{80.5}  & 98.4  & \textbf{91.6}\\
\hline
\end{tabular}}
\end{table}

\begin{table}[!htb]
      \centering
        \caption{Ablation study  on Office-31 dataset}
        \vspace{-0.2cm}
\setlength{\tabcolsep}{+0.3mm}{
\begin{tabular}{lcccccccc|c|c|c|c|c|c|c|c|}
\hline \label{tab:ab}
Task & A$\shortrightarrow$W &  A$\shortrightarrow$D & W$\shortrightarrow$A & W$\shortrightarrow$D & D$\shortrightarrow$A & D$\shortrightarrow$W  & \textbf{Ave.}\\
\hline
$CAJNet-\mathcal{A/K/DA}$	& 91.8 &	88.4 &	78.2 &	98.2 &	77.5 &	97.4 &	88.6\\
$CAJNet-\mathcal{K/DA}$ & 92.6 &	89.9 &	78.5 &	98.2&	78.9 &	97.6&	89.3	\\
$CAJNet-\mathcal{A/K}$ & 93.3 &94.4&	79.2&	98.2&	79.5&	97.4&	90.3 \\
$CAJNet-\mathcal{A/DA}$  & 93.8 &	94.6 &	79.2 &	98.5&	79.5 &	97.6&	90.5	\\
$CAJNet-\mathcal{K}$ & 93.9  & 94.6  & 79.4 &  98.7  & 79.8  & 97.9  & 90.7	\\
$CAJNet-\mathcal{DA}$ & 95.6 &	94.6 &	79.0 &	\textbf{99.2} &	78.9 &	98.0&	90.9 \\
$CAJNet-\mathcal{A}$ & 95.9 & 95.3 & 79.3 & \textbf{99.2} & 79.8 & 98.0 	& 91.3	\\
\textbf{\textit{CAJNet}}& \textbf{96.0}  & \textbf{95.6}  & \textbf{79.9} & \textbf{ 99.2 } & \textbf{80.5}  & \textbf{98.4} & \textbf{91.6}\\
\hline
\end{tabular}}
\vspace{-0.3cm}
\end{table}

\subsection{Results.}
The performance on Office + Caltech-10, Office-Home and Office-31 are shown in Tabs.~\ref{tab:OC+10}-\ref{tab:O31}. For a fair comparison, we highlight in bold those methods that are re-implemented using our extracted features, and other methods are directly reported from their original papers. Our \textit{CAJNet} model outperforms all state-of-the-art methods in terms of average accuracy (especially in the Office-Home dataset). It is compelling that our \textit{CAJNet} model substantially enhances the classification accuracy on difficult adaptation tasks (e.g., D$\shortrightarrow$A task in the Office-31 dataset and the challenging Office-Home dataset, which has a larger number of categories and different domains are visually dissimilar). 
To demonstrate the contribution of the proposed joint features, we also report the performance of marginal features alone ($CAJNet-\mathcal{C}$), conditional features alone ($CAJNet-\mathcal{M}$), and the joint features (\textit{CAJNet}) in Tab.~\ref{tab:O31}. We can find that the marginal features provide slightly higher accuracy than the proposed conditional features. This is because marginal features have more information than conditional features ($d >> C$). However, the conditional features are still important in improving performance since \textit{CAJNet} with joint features always has the highest accuracy, especially in the difficult Office-Home dataset. These experiments demonstrate the ability of the \textit{CAJNet} model to align the marginal and conditional distributions of two domains. 
\begin{figure}[b]
    \centering
    \includegraphics[width=0.48\textwidth]{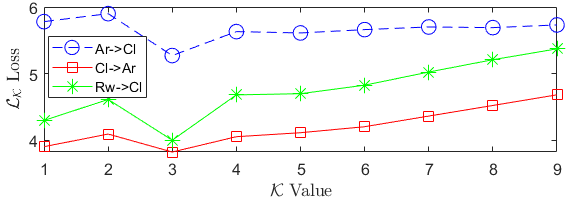}
    \caption{Effect of different $\mathcal{K}$ values on $\mathcal{L_K}$ loss.}
    \label{fig:loss}
\end{figure}
\section{Discussion}
There are two compelling advantages of the \textit{CAJNet} model. First, we compute the joint features that minimize the joint discrepancy of two domains and align the joint distribution simultaneously. Secondly, we adversarially learn the domain invariant features, while the proposed top-$\mathcal{K}$ label further updated the predicted labels via the favor of its neighbors.  However, we observe that our model is compromised in some tasks (W$\shortrightarrow$C in Office + Caltech-10 and D$\shortrightarrow$W in Office-31 dataset) 
and so we cannot guarantee that our model always beats all other methods.  We also conduct top-$\mathcal{K}$ label analysis and perform an ablation study to determine the impact of different loss functions using Office-31 dataset. Other datasets show the same effect and hence are ignored.

\paragraph*{Top-$\mathcal{K}$ label analysis}\label{sec:klabel}
$\mathcal{K}$ is the only hyperparameter in  \textit{CAJNet}, controlling how many neighbors will be used for updated labels.  Fig.~\ref{fig:loss} shows the impact of $\mathcal{K}$ on 
$\mathcal{L_K}$ loss using three tasks in the Office-Home dataset. 
We find that $\mathcal{L_K}$ achieves the smallest value when $\mathcal{K}=3$.
$\mathcal{L_K}$ increases when $\mathcal{K} > 4$ since the further away, the more likely noise labels will be included. Other tasks show the same effect.
Therefore, the top-$\mathcal{K}$ label is effective in updating predictions if correlated labels are not in the same class.

\paragraph*{Ablation study}\label{sec:abl}
To better demonstrate the effects of different loss functions on classification accuracy, we present an ablation study in Tab.~\ref{tab:ab} ($\mathcal{A}$: adversarial domain loss, $\mathcal{K}$: top-$\mathcal{K}$ correlated loss, and $\mathcal{DA}$: distribution alignment loss). Notice that, $\mathcal{L_S}$ is required during the training, and cannot be eliminated. ``$CAJNet-\mathcal{A/K/DA}$” is implemented without adversarial domain loss, top-$\mathcal{K}$ correlated loss, and distribution alignment loss. It is a simple model, which only reduces the source risk without minimizing the domain discrepancy. 
With the increasing of the number of loss functions, the robustness of our model keeps improving. 
Therefore, the proposed top-$\mathcal{K}$ approach is effective in improving performance, and we can conclude that all of the loss functions are helpful and important in minimizing the target domain risk.

\section{Conclusion}
We have proposed a correlated adversarial joint discrepancy adaptation network method (\textit{CAJNet}) to overcome prior limitations in tuning parameters and aligning joint distributions of two domains by minimizing a four component loss function. The defined top-$\mathcal{K}$ labels can help to further correct predicted labels of the two domains. Explicit domain-invariant features are learned through such a cross-domain training scheme. Experiments on three benchmark datasets show the robustness of our proposed \textit{CAJNet} model.

\small
\bibliographystyle{unsrt}
\bibliography{mybibliography}

\end{document}